\documentclass{article}
\usepackage{spconf,amsmath,graphicx}
\usepackage[ruled,vlined]{algorithm2e}

\newcommand{\Trm}{\mathrm{T}}

\newcommand{\wbf}{\mathbf{w}}
\newcommand{\xbf}{\mathbf{x}}
\newcommand{\ybf}{\mathbf{y}}
\newcommand{\zbf}{\mathbf{z}}

\newcommand{\Ibf}{\mathbf{I}}

\newcommand{\Rbf}{\mathbf{R}}
\newcommand{\Sbf}{\mathbf{S}}

\newcommand{\Zbf}{\mathbf{Z}}

\newcommand{\Bcal}{\mathcal{B}}
\newcommand{\Ucal}{\mathcal{U}}

\newcommand{\Scal}{\mathcal{S}}

\newcommand{\Ncal}{\mathcal{N}}

\newcommand{\Acal}{\mathcal{A}}

\newcommand{\Jcal}{\mathcal{J}}

\usepackage[caption=false]{subfig}

\usepackage{amsfonts}
\usepackage{cite}
\usepackage{flushend}
\usepackage{balance}

\title{Communication-Efficient Online Federated Learning Framework for Nonlinear Regression}
\name{Vinay Chakravarthi Gogineni$^{\star}$, Stefan Werner$^{\star}$, Yih-Fang Huang$^{\dagger}$, Anthony Kuh$^{\ddagger}$\thanks{The Research Council of Norway supported this work.}}
\address{$^{\star}$Dept. of Electronic Systems, Norwegian University of Science and Technology-NTNU, Norway\\ $^{\dagger}$Dept. of Electrical Engineering, University of Notre Dame, Notre Dame, IN, USA\\ $^{\ddagger}$Dept. of Electrical and Computer Engineering, University of Hawaii, Hawaii, USA\\
E-mails: \{vinay.gogineni, stefan.werner\}@ntnu.no, huang@nd.edu, kuh@hawaii.edu}
\begin{document}
\ninept
\maketitle
\begin{abstract}
Federated learning (FL) literature typically assumes that each client has a fixed amount of data, which is unrealistic in many practical applications. Some recent works introduced a framework for online FL (Online-Fed) wherein clients perform model learning on streaming data and communicate the model to the server; however, they do not address the associated communication overhead. As a solution, this paper presents a partial-sharing-based online federated learning framework (PSO-Fed) that enables clients to update their local models using continuous streaming data and share only portions of those updated models with the server. During a global iteration of PSO-Fed, non-participant clients have the privilege to update their local models with new data. Here, we consider a global task of kernel regression, where clients use a random Fourier features-based kernel LMS on their data for local learning. We examine the mean convergence of the PSO-Fed for kernel regression. Experimental results show that PSO-Fed can achieve competitive performance with a significantly lower communication overhead than Online-Fed.

\end{abstract}
\section{Introduction}
\label{sec:intro}
Federated learning (FL) \cite{FL_acm, FL_spm, FL_tcom} has emerged as an appealing distributed learning framework that allows a network of edge devices to train a global model without revealing local data to others. Among the features that make federated learning (FL) stand out from typical distributed learning, the four most significant are as follows. First, the local data of each client device are not independent and identically distributed (i.e., non-IID) and unbalanced in the amount  \cite{FL_niid1, FL_niid2}. Besides that, devices will not disclose the local data to the server or any other client during the training process. Second, federated learning was originally designed to use data collected on battery-constrained or low-performance devices, which also tend to have a low memory capacity \cite{FL_mem}. The memory should, therefore, not be depleted by local learning. Third, clients frequently go offline or have limited bandwidth or expensive connections \cite{FL_asyn, FL_channel}. Lastly, the reliability of clients in federated learning is questionable \cite{FL_privacy1, FL_privacy2}. Clients with malicious intent may attempt to degrade the global model's reliability. In this paper, we are primarily concerned with reducing communication overhead. 

One of the most popular FL methods is federated averaging (FedAvg) \cite{FedAvg}. The workflow of FedAvg is as follows: At the start of each global round, a random fraction of clients receive a copy of the global model from the server (generally, clients are selected uniformly, but various methods may be used \cite{FedCS, FedNu, ISFedAvg}). Using the global model and local data, the selected client then performs multiple iterations of local learning and sends the updated local model to the server. The server then aggregates these updated local models to build a new global model, and the process repeats. Even though FedAvg minimizes communication overhead by executing multiple local updates at each selected client before communicating to the server, its learning accuracy largely depends on the number of epochs (a critical factor in deciding the communication interval) performed at each client \cite{FedAvg, FedProx}. 

As we can see from the workflow of FL, the participating clients have to communicate the model back and forth with the server in each global iteration. Furthermore, the FL framework generally takes hundreds or thousands of iterations to finalize the global model, and the size of a modern machine learning model is on the order of a billion. The FL framework would have to deal with this enormous communication overhead if utilized. Various solutions have been explored in the literature for reducing the communication overhead associated with FL. Among these, communication-mitigated federated learning (CMFL) \cite{CMFL} discards the irrelevant updates from clients by checking the alignment between the local and global updates tendency.  A couple of procedures have been proposed in \cite{FL_ssupdates} for reducing uplink communication overhead. The first one is the structured update: clients update the model in a restricted space parametrized with fewer variables. The other one is sketch update: clients update a full model and then compress it using a combination of $1$-bit quantization, random rotations, and subsampling before sending it to the server. While sketch updates reduce communication overhead, they are time-consuming and incur additional complexity for clients. By discarding unimportant client updates for global learning (i.e., when numerous model parameters remain unchanged), structured communication reduction for federated learning (FedSCR) \cite{FedSCR} can reduce the communication overhead.

The FL approaches outlined above assume a fixed amount of training data at each client, which is impractical in many real-life scenarios, e.g., in wireless communications. Instead, clients may have access to new data or a stream of data during the training \cite{MLstreamingdata}. Recently, in \cite{O-Fed}, the concept of online federated learning (Online-Fed) is discussed; however, no concrete mathematical equations have been given. In Online-Fed, the clients perform online learning on a continuous stream of local data while the server aggregates model parameters received from the clients. On the other hand, the asynchronous online federated learning framework (ASO-Fed) \cite{FL_asyn} focused mainly on learning a global model from asynchronous client updates. However, these frameworks have not addressed the communication overhead associated with them. 

In this paper, we present an energy-efficient online federated learning framework, namely, partial-sharing-based online federated learning (PSO-Fed), wherein each client updates its local model using continuous streaming data and then shares just a portion of the updated model parameters with the server. In contrast to Online-Fed, PSO-Fed permits non-participant clients to update their local models when they access new data during global iteration. To demonstrate the efficacy of PSO-Fed, we considered nonlinear regression in a non-IID setting. For this, we employ random Fourier features-based kernel LMS (RFF-KLMS) \cite{klms, rff, rffklms} to perform the nonlinear regression task locally at each client. In addition, mean convergence analysis of PSO-Fed is provided for these settings. Finally, we perform numerical experiments on synthetic non-IID data, and our results confirm that PSO-Fed achieves competitive performance at very low communication overhead compared to Online-Fed.

\section{Problem Formulation and Algorithm Description}
\label{sec:proalgo}
In this section, we first introduce the Online-Fed in the context of kernel regression. Then, we present an energy-efficient version called PSO-Fed. In the following, we consider a scenario wherein $K$ geographically distributed clients communicate with a global server. At every time instance $n$, every client $k$ has access to a continuous streaming signal $x_{k, n}$ and associated desired outputs $y_{k, n}$, assumed to be described by the model:
\begin{align}\label{eq1}
    y_{k, n} = f(\xbf_{k, n}) + \nu_{k, n},
\end{align}
where $f(\cdot)$ is a continuous nonlinear model to be estimated collaboratively using clients' data, $\xbf_{k, n}=[x_{k, n}, \cdots, x_{k, n-L+1}]^{\text{T}}$ is the local data vector of size $L \times 1$, and $\nu_{k, n}$ is the observation noise. For client $k$, we then define the local optimization function for estimating $f(\cdot)$ as follows:
\begin{align}\label{eq2}
    \Jcal_k(\wbf_k) = E\left[|y_{k, n} -\hat{y}_{k, n}|^2 \right],
\end{align}
with $\hat{y}_{k, n} = \wbf_{k}^\Trm \zbf_{k, n}$, where the local model parameter vector $\wbf_k \in \mathbb{R}^D$, is a linear representation of the nonlinear model $f(\cdot)$ in a random Fourier feature (RFF) space of dimension $D$, and $\zbf_{k, n} \in \mathbb{R}^D$ being the mapping of $\xbf_{k, n}$ into RFF space. Cosine, exponential, and Gaussian feature functions \cite{rff, rffklms} can be used to represent $\xbf_{k, n}$ in the RFF space. Then, the optimization at the global server is 
\begin{align}\label{eq3}
    \Jcal(\wbf) = \frac{1}{K} \sum\limits_{k=1}^{K} \Jcal_k(\wbf).
\end{align}
Here, the goal is to find an estimate of global optimal representation of the function $f(\cdot)$ in RFF space, i.e., $\wbf^{\star}$ as:
\begin{align}\label{eq4}
    \wbf_n= \min_{{\wbf}}^{} \Jcal(\wbf).
\end{align}
\subsection{Online-Fed}
In each global iteration $n$, the server selects a subset of clients and share the global model $\wbf_n$ with them. Thereafter, the selected clients $\forall k \in \Scal_n$ ($\Scal_n$ is a set containing selected client indices in global iteration $n$) run a stochastic gradient descent to solve the local optimization problem $\Jcal_k(\wbf_k)$ as follows:
\begin{align}\label{eq5}
\wbf_{k, n+1}= \wbf_{n} + \mu \hspace{1mm} \zbf_{k, n} \hspace{1mm} \epsilon_{k, n},
\end{align}
where $\mu$ is the learning rate and $\epsilon_{k, n}= y_{k, n}- \wbf_{n}^\Trm \zbf_{k, n}$. These clients communicate the updated local models to the server. Then, the server aggregates the received updates as 
\begin{align}\label{eq6}
    \wbf_{n+1} = \frac{1}{|\Scal_n|} \sum\limits_{k \in \Scal_n }^{} \wbf_{k, n+1},
\end{align}
where $|\Scal_n|$ denotes the cardinality of $\Scal_n$. We see from the workflow of Online-Fed that clients not selected during the $n$th global iteration do not perform a local model update, despite having access to the local streaming data. Regardless of whether they update, whenever the global server selects them, the latest local model will be replaced by the global model without considering the last update made locally. This issue hinders the performance.  Furthermore, the amount of communication taking place at each global iteration is still significant. One solution is to use the concept of structure updates or sketch updates \cite{FL_ssupdates}. Our solution to this problem stems from a different approach, namely, partial-sharing concepts \cite{PdLMS, PdRLS} that are very attractive for communication-efficient distributed learning. 
\subsection{PSO-Fed}
In the proposed partial-sharing-based online federated learning (PSO-Fed), clients and the server exchange just a fraction of their model parameters in each update round. In order to keep track of the model parameters being exchanged in each communication round, the client and server maintain selection matrices.  

To this end, at every global iteration $n$, the model parameters to be exchanged between clients and the server are specified by a diagonal selection matrix  ${\bf S}_{k, n}$ of size $D \times D$. On its principal diagonal, ${\bf S}_{k, n}$ contains $M$ ones and $D-M$ zeros. In $\Sbf_{k, n}$, the positions of ones specify which local model parameters to be exchanged with the server. As in \cite{PdLMS, PdRLS}, we can select the  $M$ model parameters either stochastically or sequentially. To simplify the implementation, we consider coordinated and uncoordinated partial-sharing. The server assigns the same initial selection matrices to all clients in coordinated partial-sharing (i.e., ${\bf S}_{1, 0}={\bf S}_{2, 0}= \cdots= {\bf S}_{K, 0}={\bf S}_0$). As a result, all clients exchange the same portion of their local model parameters with the server. On the other hand, in uncoordinated partial-sharing, the server assigns initial selection matrices randomly to clients (i.e., ${\bf S}_{1, 0}\neq{\bf S}_{2, 0} \neq \cdots \neq {\bf S}_{K, 0}$). Both schemes belong to sequential and stochastic partial-sharing families, respectively. For the current global iteration $n$, the entry selection matrix ${\bf S}_{k, n}$ can be obtained via a right circular shift of ${\bf S}_{k, n-1}$. In this process, each entry will be exchanged $M$ times over $D$ iterations, so the probability of a specified model parameter being exchanged with the server is $\frac{M}{D}$. With the help of selection matrices, Online-Fed workflow can be alternatively expressed as:
\begin{subequations}\label{eq7}
\begin{align}
 \wbf_{k, n+1}= \Sbf_{k, n} \wbf_{n} +  (\Ibf_D -\Sbf_{k,n}) \wbf_{n} + \mu \hspace{1mm} \zbf_{k, n} \hspace{1mm} \epsilon_{k, n},   
\end{align}
with $\epsilon_{k, n}= y_{k, n}- (\Sbf_{k, n} \wbf_{n} +  (\Ibf_D -\Sbf_{k, n}) \wbf_{n})^\Trm \zbf_{k, n}$
\begin{align}
    \wbf_{n+1} = \frac{1}{|\Scal_n|} \sum\limits_{k \in \Scal_n }^{} \Sbf_{k, n+1} \wbf_{k, n+1} +  (\Ibf_D -\Sbf_{k, n+1}) \wbf_{k, n+1}).
\end{align}
\end{subequations}

Since PSO-Fed limits the exchange of model parameters, the server does not have access to all participating clients' model parameters during the aggregation phase.  Similarly, the participating clients do not have access to entire global model parameters; therefore, they will use their previous model parameters in place of the unknown portions. Participating clients use $(\Ibf_D -\Sbf_{k,n}) \wbf_{k, n}$ in place of $(\Ibf_D -\Sbf_{k,n}) \wbf_{n}$ and the server uses $(\Ibf_D -\Sbf_{k, n+1}) \wbf_{n}$ in place of $(\Ibf_D -\Sbf_{k, n+1}) \wbf_{k, n+1}$. The non-participating clients use their previous local models to perform the local learning. The proposed PSO-Fed is summarized in Algorithm~\ref{Algo1}. 

It is important to note that even clients do not take part in all global iterations, the PSO-Fed still permits them to perform local updates as long as they have access to new data. In contrast, as touched upon above, state-of-the-art approaches replace local models with the global model whenever clients are selected for contributing to the model update, making local updates during communication-dormant times futile. It is evident that clients have better control over local learning with PSO-Fed than with current state-of-the-art FL approaches.  
\begin{algorithm}[t!]
\DontPrintSemicolon 
\textbf{Initialization}: global model $\wbf_0$, local model $\wbf_{k, 0}$, RFF space dimension $D$  and selection matrices ${\Sbf}_{k, 0}$, $\forall k \in \Scal$,\\[1mm]
\textbf{For} \,$n=1$ to $N$ \\[1mm]
The server randomly selects a subset $\Scal_n$ of $K$ clients and communicate $\Sbf_{k, n} {\bf w}_{n}$ to them, \\[1mm]
\textbf{Client Local Update}:\\[1mm]
\hspace{5mm}\textbf{If} \,$k \in \Scal_n$\\[-6mm]
\begin{align*}
\begin{split}
\epsilon_{k, n}&= y_{k, n}- (\Sbf_{k,, n} \wbf_{n} +  (\Ibf_D -\Sbf_{k, n}) \wbf_{k, n})^\Trm \zbf_{k, n}, \\
\wbf_{k, n+1}&= \Sbf_{k,, n} \wbf_{n} +  (\Ibf_D -\Sbf_{k,n}) \wbf_{k, n} + \mu \hspace{1mm} \zbf_{k, n} \hspace{1mm} \epsilon_{k, n}, 
\end{split}
\end{align*}
\hspace{5mm}\textbf{Else} \\[-7mm]
\begin{align*}
\begin{split}
\epsilon_{k, n}&= y_{k, n}- \wbf_{k, n}^\Trm \zbf_{k, n}, \\
\wbf_{k, n+1}&= \wbf_{k, n} + \mu \hspace{1mm} \zbf_{k, n} \hspace{1mm} \epsilon_{k, n}, 
\end{split}
\end{align*}
\hspace{5mm}\textbf{EndIf}\\[1mm]
The clients $\forall k \in \Scal_n$ communicate ${\Sbf}_{k, n+1} \wbf_{k, n+1}$ to the server, where ${\Sbf}_{k, n+1}= \text{circshift}({\Sbf}_{k, n}, \tau)$,\\[1mm]
\textbf{Aggrigation at the Server}:\\[1mm]
\hspace{5mm}The server updates the global model as,
\begin{align*}
\begin{split}
\wbf_{n+1}= \frac{1}{|\Scal_n|}\sum_{k \in \Scal_n} {\Sbf}_{k, n+1} \wbf_{k, n+1} + ({\Ibf}_D-{\Sbf}_{k, n+1}) \wbf_{n}.
\end{split}
\end{align*}
\textbf{EndFor}\\
\caption{\texttt{\textbf{PSO-Fed}}. $K$ clients, learning rate $\mu$, set of all clients $\Scal$, and circular shift variable $\tau$.}\label{Algo1}
\end{algorithm}
\section{Convergence Analysis}
\label{sec:analysis}
In this section, we examine the mean convergence of PSO-Fed. Before proceeding to the analysis, we define the global optimal extended model parameter vector ${\bf w}^{\star}_{e} = {\bf 1}_{K+1} \otimes {\bf w}^{\star}$, extended estimated global model parameter vector ${\bf w}_{e, n}=\text{col}\{{\bf w}_{n}, \wbf_{1, n}, \ldots, \newline \wbf_{K, n}\}$, extended input data matrix ${\bf Z}_{e, n}=\text{blockdiag}\{\textbf{0}, {\bf z}_{1, n}, \ldots, \newline {\bf z}_{K, n}\}$ and  extended observation noise vector $\boldsymbol{\nu}_{e, n}=\text{col}\big\{\textbf{0}, \nu_{1, n}, \newline \ldots, \upsilon_{K, n} \big\}$, where $\text{col}\{\cdot\}$ and $\text{blockdiag}\{\cdot\}$ represent column-wise stacking operator and block diagonalization operator, respectively. The symbol ${\bf 1}_{K+1}$ is a $(K+1) \times 1$ column vector with each element taking the value one. From the above definitions, we can write
\begin{equation}\label{eq8}
\begin{split}
{\bf y}_{e, n}&= \text{col}\{\textbf{0}, y_{1, n}, y_{2, n}, \ldots, y_{K, n}\}={\bf Z}^{\text{T}}_{e, n} {\bf w}_{e}^{\star} + \boldsymbol{\nu}_{e, n}, \\
\boldsymbol{\epsilon}_{e, n}&=\text{col}\big\{ \textbf{0}, \epsilon_{1, n}, \epsilon_{2, n}, \ldots, \epsilon_{K, n} \big\}= {\bf y}_{e, n} - {\bf Z}^{\text{T}}_{e, n} \boldsymbol{\Acal}_{\Sbf, n} {\bf w}_{e, n},
\end{split}
\end{equation}
with
\begin{align}\label{eq9}
\begin{split}
&\boldsymbol{\Acal}_{\Sbf, n} = \\
&\begin{bmatrix}
    \Ibf_{D} & \textbf{0} & \textbf{0} & \dots  & \textbf{0} \\
    a_{1, n} \Sbf_{1, n} & \Ibf_{D} - a_{1, n} \Sbf_{1, n} & \textbf{0} & \dots  & \textbf{0} \\
    \vdots & \vdots & \vdots & \ddots & \vdots \\
    a_{K, n} \Sbf_{K, n} & \textbf{0} & \textbf{0} & \dots  & \Ibf_{D} - a_{K, n} \Sbf_{K, n} 
\end{bmatrix},  
\end{split}
\end{align}
where $a_{k, n}=1$ if the client $k \in \Scal_n$, and zero otherwise. Using these definitions, the global recursion of PSO-Fed can be stated as
\begin{equation}\label{eq10}
\begin{split}
{\bf w}_{e, n+1}&= \boldsymbol{\mathcal{B}}_{\Sbf, n+1} \big(\boldsymbol{\mathcal{A}}_{\Sbf, n} {\bf w}_{e, n}+ \mu \hspace{1mm} {\bf Z}_{e, n} \hspace{1mm}  \boldsymbol{\epsilon}_{e, n} \big),
\end{split}
\end{equation}
where 
\begin{align}\label{eq11}
\begin{split}
&\boldsymbol{\Bcal}_{\Sbf, n+1} =  \\
&\begin{bmatrix}
    \Ibf_{D} - \hspace{-1mm} \sum\limits_{k \in \Scal_{n}}^{} \hspace{-3mm} \frac{a_{k, n}}{|\Scal_n|}  \Sbf_{k, n+1}& \frac{a_{1, n}}{|\Scal_n|}  \Sbf_{1, n+1}  & \dots  & \frac{a_{K, n}}{|\Scal_n|}  \Sbf_{K, n+1} \\
    \textbf{0} & \Ibf_{D}  & \dots  & \textbf{0} \\
    \vdots & \vdots  & \ddots & \vdots \\
    \textbf{0} & \textbf{0} & \dots  & \Ibf_{D}  
\end{bmatrix}. 
\end{split}
\end{align}

We make the following assumptions to establish the convergence condition for PSO-Fed:\newline
\textbf{A1}: At each client $k$, the input signal vector ${\bf z}_{k,n}$ is drawn from a wide-sense stationary multivariate random sequence with correlation matrix ${\bf R}_{k}= {\rm E}[{\bf z}_{k,n} {\bf z}^\Trm_{k,n}]$. \newline
\textbf{A2}: The noise process $\nu_{k, n}$ is assumed to be zero-mean i.i.d. and independent of all input and output data,  \newline
\textbf{A3}: At each client $k$, the model parameter vector is taken to be independent of input signal vector. \newline
\textbf{A4}: The selection matrices $\mathbf{S}_{k, n}$ are assumed to be independent of any other data; in addition, $\mathbf{S}_{k, n}$ and $\mathbf{S}_{l, m}$ are independent, for all $k\neq l$ and $m \neq n$.

Denoting $\widetilde{\mathbf{w}}_{e,n}={\bf w}_{e}^{\star}-{\bf w}_{e,n}$, and utilizing the fact that ${\bf w}_{e}^{\star} = \boldsymbol{\Bcal}_{\Sbf, n+1}  \boldsymbol{\Acal}_{\Sbf, n} {\bf w}_{e}^{\star}$ (since $\boldsymbol{\Bcal}_{\Sbf, n+1} {\bf w}_{e}^{\star} =  \boldsymbol{\Acal}_{\Sbf, n} {\bf w}_{e}^{\star} = {\bf w}_{e}^{\star}$,  one can easily prove this result), then from \eqref{eq10}, $\widetilde{\mathbf{w}}_{e, n+1}$ can be recursively expressed as
\begin{equation}\label{eq12}
\begin{split}
\widetilde{{\bf w}}_{e, n+1}=& \boldsymbol{\Bcal}_{\Sbf, n+1} \big(\mathbf{I} -\mu {\bf Z}_{e, n} {\bf Z}_{e, n}^\text{T} \big) \boldsymbol{\Acal}_{\Sbf, n} \widetilde{{\bf w}}_{e, n}\\
&-\mu \boldsymbol{\Bcal}_{\Sbf, n+1} {\bf Z}_{e, n} \boldsymbol{\nu}_{e,n}.
\end{split}
\end{equation}
Applying expectation $\mathbb{E}[\cdot]$ on both sides of \eqref{eq12} and using assumptions $\textbf{A1}-\textbf{A4}$, we obtain 
\begin{equation}\label{eq13}
\begin{split}
 \mathbb{E}[\widetilde{{\bf w}}_{e, n+1}]= \mathbb{E}[\boldsymbol{\Bcal}_{\Sbf, n+1}] \big( \mathbf{I} - \mu \boldsymbol{\mathcal{R}}_{e}\big) \mathbb{E}[\boldsymbol{\Acal}_{\Sbf, n}]  \mathbb{E}[\widetilde{{\bf w}}_{e, n}],
\end{split}
\end{equation}
where $\boldsymbol{\mathcal{R}}_{e}=\text{blockdiag}\{ \textbf{0}, \mathbf{R}_{1}, \mathbf{R}_{2}, \ldots, \mathbf{R}_{K}\}$. One can see that $\text{E}\big[\widetilde{{\bf w}}_{e, n}\big]$ converges under $\| \mathbb{E}[\boldsymbol{\Bcal}_{\Sbf, n+1}] \big( \mathbf{I} - \mu \boldsymbol{\mathcal{R}}_{e}\big) \mathbb{E}[\boldsymbol{\Acal}_{\Sbf, n}]\| < 1$ for every $n$, where $\| \cdot\|$ is any matrix norm. Since $\| \mathbb{E}[\boldsymbol{\Bcal}_{\Sbf, n+1}]\| = 1 $ and $\| \mathbb{E}[\boldsymbol{\Acal}_{\Sbf, n}]\| = 1$, the above convergence condition reduces to $\|\mathbf{I} - \mu \boldsymbol{\mathcal{R}}_{e}\|  < 1$, or, equivalently, $\forall k,i : |1 - \mu \lambda_i(\Rbf_{k})| <1$, where $\lambda_i(\cdot)$ is the $i$th eigenvalue of its argument matrix. After solving the above convergence condition, we finally have  following first-order convergence condition:
\begin{equation}\label{eq14}
\begin{split}
0 < \mu < \dfrac{2}{\max\limits_{\forall k} \{ \max\limits_{\forall i} \{ \lambda_i (\Rbf_{k}) \}  \}  }.
\end{split}
\end{equation}
\section{Numerical Simulations}
\label{sec:simulations}
\begin{figure}[t!]
\centering
\subfloat[\label{PSO-Fed-cor}]{{\includegraphics[width=.425\textwidth, height =68mm]{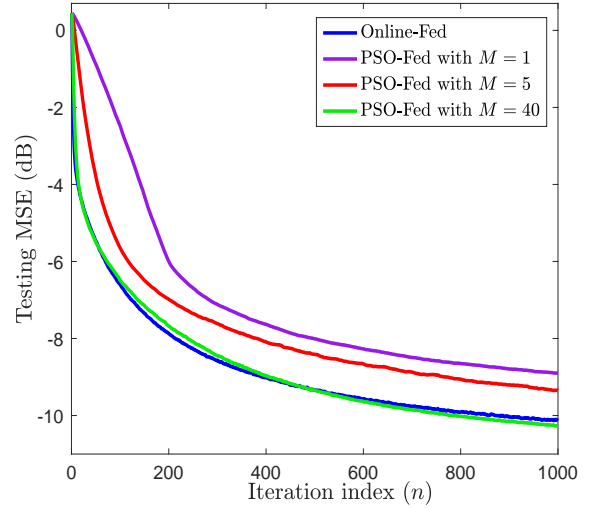} }}\\[-0.1mm]
\subfloat[\label{PSO-Fed-uncor}]{{\includegraphics[width=.425\textwidth, height =68mm]{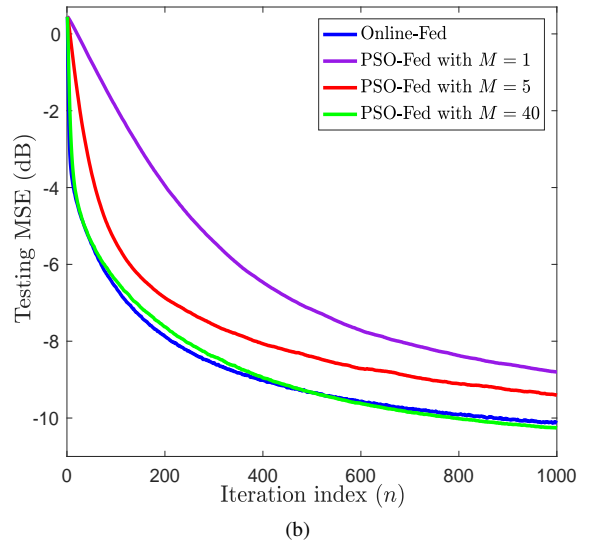} }}
\caption{Performance of PSO-Fed: (a). Coordinated partial-sharing. (b). Uncoordinated partial-sharing.}
\label{fig:PSO-Fed}
\vspace{-5mm}
\end{figure}

In this section, experimental results are presented to examine the performance of PSO-Fed. Our experiment considers $K = 100$ clients with access to a global server. At every client $k$, synthetic non-IID input signal $x_{k, n}$ and corresponding observed output are generated so that they are related via the following model:
\begin{align}
   f(\xbf_{k, n}) &= \sqrt{x^2_{k,1, n} + \sin^2(\pi \hspace{1mm} x_{k, 4, n} )} \nonumber \\
 &\hspace{5mm} + \big(0.8 - 0.5\exp(-x^2_{k, 2, n} \big) x_{k, 3, n} + \nu_{k, n}.
\end{align}
The input signal at each client $x_{k, n}$ was generated by driving a first-order autoregressive (AR) model: $x_{k, n}= \theta_{k} \hspace{1mm} x_{k, n-1} + \sqrt{1- \theta_k^{2}} \hspace{1mm} u_{k, n}$, $\theta_k \in \Ucal(0.2, 0.9)$, where $u_{k, n}$ was drawn from a Gaussian distribution $\Ncal(\mu_k, \sigma_{u_k}^2)$, with $\mu_k \in \Ucal(-0.2, 0.2)$ and $\sigma_{u_k}^2 \in \Ucal(0.2, 1.2)$, respectively (where $\Ucal(\cdot)$ indicates the uniform distribution). The observation noise $\nu_{k, n}$ was taken as zero mean i.i.d. Gaussian with variance $\sigma_{\nu_k}^2 \in \Ucal(0.005, 0.03)$. Using a Cosine feature function, $x_{k,n}$  was mapped into the RFF space  whose dimension was fixed to $200$. All simulated algorithms were set to the same learning rate of $0.75$ for each client. The server implemented uniform random selection procedure to select $|\Scal_n|=4$ clients in every global iteration $n$. By calculating the average mean-square error (MSE) on test data after each global iteration $n$, we evaluated the simulation performance:
\begin{align}
\text{MSE} = \frac{1}{N_{\text{test}}}\|\ybf_{\text{test}} -  \Zbf_{\text{test}}^{\Trm} \wbf_n\|^2_2, 
\end{align}
where $\{\Zbf_{\text{test}}, \ybf_{\text{test}} \}$ is the test data set ($N_{\text{test}}$ examples in total) covering all clients data. In order to perform the nonlinear regression task, the proposed PSO-Fed was simulated for a variety of values of $M$ (number of model parameters exchanged between the server and clients). In addition, we also simulated the Online-Fed for comparative evaluation. The learning curves (i.e., test MSE in dB against the global iteration index $n$) are obtained by averaging over $500$ independent experiments, are shown in Figs.~\ref{PSO-Fed-cor} and \ref{PSO-Fed-uncor} for coordinated and uncoordinated partial-sharing schemes, respectively. From Fig.~\ref{fig:PSO-Fed}, the following interesting observations can be made:
\begin{enumerate}
\item With PSO-Fed, we can achieve competitive results at a lower communication cost than Online-Fed. Firstly, PSO-Fed exhibits a slower convergence rate but with a similar steady-state MSE as Online-Fed at small  values of $M$ (e.g., $1$). As $M$ increases to higher values (e.g., $5$ and $40$), its convergence becomes faster. In summary, PSO-Fed exhibits a similar convergence rate with a minor improvement in steady-state MSE when $M\geq40$. 
\item Because $M$ is much smaller than $D$, PSO-Fed's communication cost is lower than that of Online-Fed. PSO-Fed behaves the same as Online-Fed when $M = 40$ but only consumes $ \frac{1}{5} $ of its communication load. As non-participating clients make updates locally if they have access to new data, partial-sharing alters only part of these locally updated models when they get a chance to communicate with the server. Thus, resulting in improved performance and reduced communication load. It is worth noting that the proposed partial-sharing does not incur any additional computational overhead, unlike sketch updates proposed in \cite{FL_ssupdates}. Keeping track of partially-shared parameter indices just requires a little extra memory.
\item Coordinated partial-sharing has a faster initial convergence speed than uncoordinated one, for very small values of $M$ (e.g., $1$ in our experiment). In particular, the coordinated scheme preserves the connectedness of clients by allowing the server to aggregate the same entries of the local model parameter vectors. However, both schemes are equally effective for large values of $M$ (e.g., $\geq 5$ in our experiment).
\end{enumerate}

\section{Conclusions}
\label{sec:conclusions}
An energy-efficient framework has been developed for online FL, called PSO-Fed. In PSO-Fed, participating clients exchange a fraction of model parameters with the server, but non-participating clients update their local model if they have access to new data. Thus, the negative effects of partial-sharing have been compensated. PSO-Fed's performance has been demonstrated via kernel regression. The convergence of PSO-Fed has been analyzed under these settings. Simulation results have shown that both coordinated and non-coordinated PSO-Fed algorithms exhibit competitive estimation performance while reducing communication costs compared to Online-Fed.

\pagebreak

\bibliographystyle{IEEEbib}

\end{document}